\documentclass[acmlarge]{acmart}
\newcommand{\final}{1}

\usepackage{wrapfig}
\usepackage{color}
\usepackage{ifthen}
\usepackage{float}
\usepackage{alltt}
\usepackage{amsmath}
\usepackage{amsthm}
\usepackage{newlfont}
\usepackage{floatflt}
\usepackage{wrapfig}
\usepackage{fixltx2e}
\usepackage{subfig}
\usepackage{multirow}
\usepackage{booktabs}
\usepackage{cleveref}
\usepackage{algorithmic}
\usepackage[export]{adjustbox}
\usepackage{mathtools, cuted}
\usepackage{CJKutf8}
\usepackage[ruled]{algorithm2e}
\setcitestyle{square}

\SetAlFnt{\small}
\SetAlCapFnt{\small}
\SetAlCapNameFnt{\small}
\SetAlCapHSkip{0pt}
\IncMargin{-\parindent}
\newcommand{\Caption}[2]{\caption[#1]{{\em #1} #2}}
\let\oldcaption\caption
\renewcommand{\caption}[2][]{\oldcaption[#1]{{\em #1} #2}}

\definecolor{figred}{rgb}{1,0,0}
\definecolor{figgreen}{rgb}{0,0.6,0}
\definecolor{figblue}{rgb}{0,0,1}
\definecolor{figpink}{rgb}{1,0.63,0.63}

\newcommand{\pseudocode}{Algorithm}
\floatstyle{plain}
\newfloat{algorithm}{tbhp}{lop}
\floatname{algorithm}{\pseudocode}

\newcommand{\filename}[1]{\url{#1}}
\newcommand{\foldername}[1]{\url{#1}}

\let\oldparagraph\paragraph
\iffalse

\renewcommand{\paragraph}[1]{\oldparagraph{\textbf{#1}.}}
\else

\renewcommand{\paragraph}[1]{\oldparagraph{{#1}.}}
\fi

\ifdefined\email
\else
\newcommand{\email}[1]{\url{#1}}
\fi

\newcommand{\revise}[1]{{\color{blue} #1}}

\ifthenelse{\equal{\final}{1}} {
    \renewcommand{\revise}[1]{{#1}}
}

\newcommand{\nothing}[1]{{}}
\acmJournal{TAP}
\acmYear{2024}
\acmVolume{21}
\acmNumber{4}
\acmArticle{14}
\acmMonth{11}
\acmDOI{10.1145/3694969}

\setcopyright{acmcopyright}

\ccsdesc[500]{Computing methodologies~Artificial intelligence}
\ccsdesc[500]{Computing methodologies~Perception}

\keywords{Human Visual Attention, Perceptual Computer Graphics, Controllable Image Generation}
\newcommand{\methodName}{GazeFusion\xspace}

\newcommand{\conditionUNCOND}{\textbf{TEXT}\xspace}
\newcommand{\conditionBBOX}{\textbf{BBOX}\xspace}
\newcommand{\conditionSMAP}{\textbf{OURS}\xspace}

\newcommand{\groupTEST}{\textbf{TEST}\xspace}
\newcommand{\groupCONTROL}{\textbf{CONTROL}\xspace}

\newcommand{\saliencySpatial}{S}
\newcommand{\saliencySpatialSeq}{\mathbf{S}}
\newcommand{\saliencyVideo}{V}
\newcommand{\saliencyVideoSeq}{\mathbf{V}}

\newcommand{\timeDuration}{T}

\newcommand{\gaussian}{G}
\newcommand{\gaussianMean}{\mu}
\newcommand{\gaussianCov}{\Sigma}
\newcommand{\gaussianWeight}{w}

\newcommand{\textPrompt}{p}

\title[GazeFusion: Saliency-Guided Image Generation]{GazeFusion: Saliency-guided Image Generation}

\author{Yunxiang Zhang}
\orcid{0000-0003-0189-1776}
\email{yunxiang.zhang@nyu.edu}
\affiliation{
 \institution{New York University}
 \country{USA}}
 
\author{Nan Wu}
\orcid{0009-0008-0866-334X}
\email{wunan@stanford.edu}
\author{Connor Z. Lin}
\orcid{0000-0003-1388-6974}
\email{connorzl@stanford.edu}
\author{Gordon Wetzstein}
\orcid{0000-0002-9243-6885}
\email{gordon.wetzstein@stanford.edu}
\affiliation{
 \institution{Stanford University}
 \country{USA}}
 
\author{Qi Sun}
\orcid{0000-0002-3094-5844}
\email{qisun@nyu.edu}
\affiliation{
 \institution{New York University}
 \country{USA}}
\begin{abstract}
\revise{Diffusion models offer unprecedented image generation power given just a text prompt. While emerging approaches for controlling diffusion models have enabled users to specify the desired spatial layouts of the generated content, they cannot predict or control where viewers will pay more attention due to the complexity of human vision. Recognizing the significance of attention-controllable image generation in practical applications, we present a saliency-guided framework to incorporate the data priors of human visual attention mechanisms into the generation process.} Given a user-specified viewer attention distribution, our control module conditions a diffusion model to generate images that attract viewers' attention toward the desired regions. To assess the efficacy of our approach, we performed an eye-tracked user study and a large-scale model-based saliency analysis. The results evidence that both the cross-user eye gaze distributions and the saliency models' predictions align with the desired attention distributions. Lastly, we outline several applications, including interactive design of saliency guidance, attention suppression in unwanted regions, and adaptive generation for varied display/viewing conditions.
\end{abstract}

\begin{document}
\maketitle
\section{Introduction}
\label{sec:intro}

The recent emergence of generative artificial intelligence marks a paradigm shift for computer graphics. Diffusion models, in particular, enable the generation and editing of both photorealistic and stylized images, videos, and 3D objects with little more than a text prompt or high-level user guidance as inputs~\cite{po2023state}. In many applications, including graphic design and advertisement, it is desirable to generate visual content that attracts a viewer's attention toward specific areas of interest. Such a human-centric control mechanism for the generation process, however, is not supported by current image diffusion models.

\begin{figure}[t]
\centering
\subfloat[saliency-guided image generation for triggering desired viewer attention]{
    \label{fig:teaser-all}
    \includegraphics[width=0.72\textwidth]{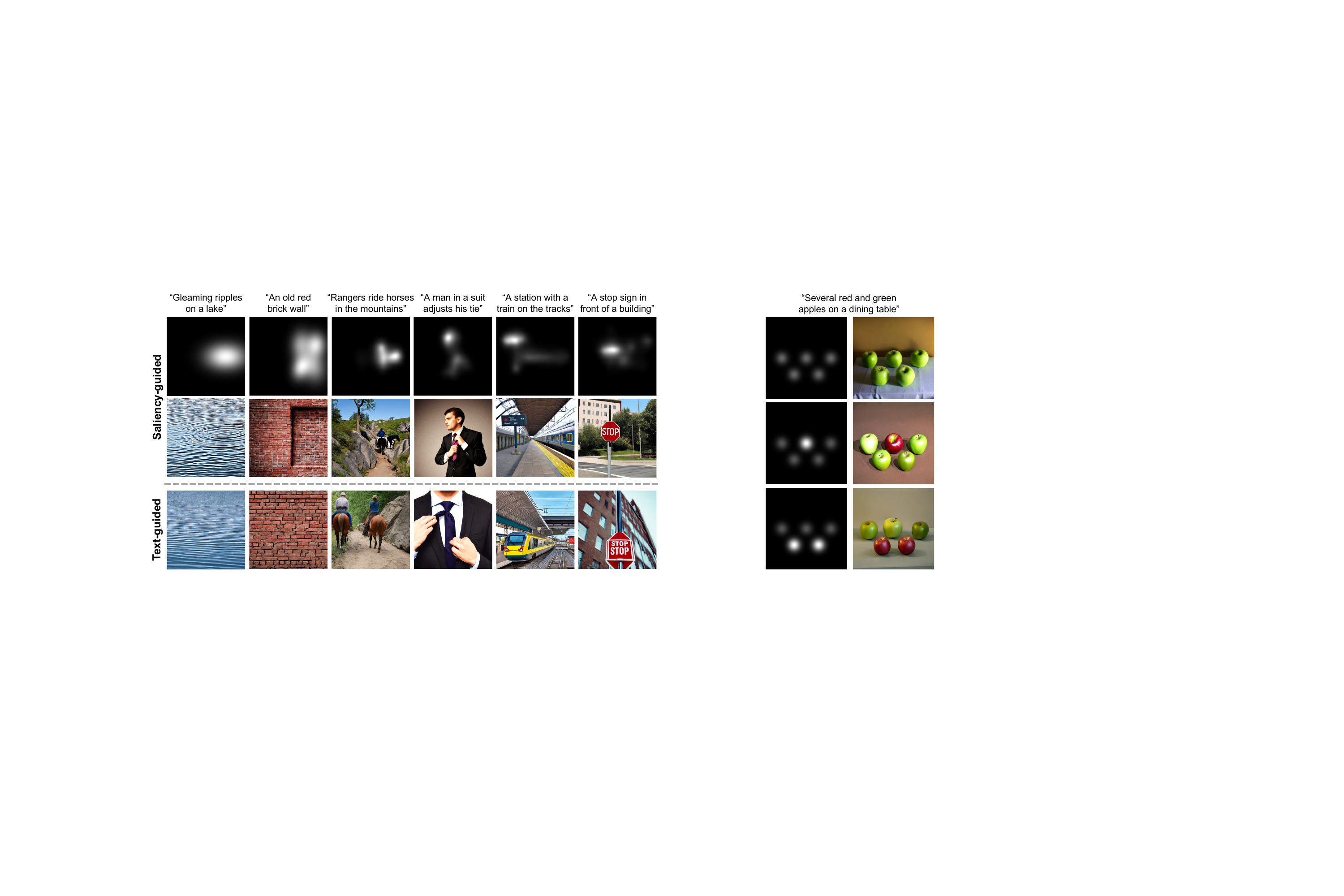}
  }
\subfloat[variations of a fixed layout]{
    \label{fig:teaser-apple}
    \includegraphics[width=0.2435\textwidth]{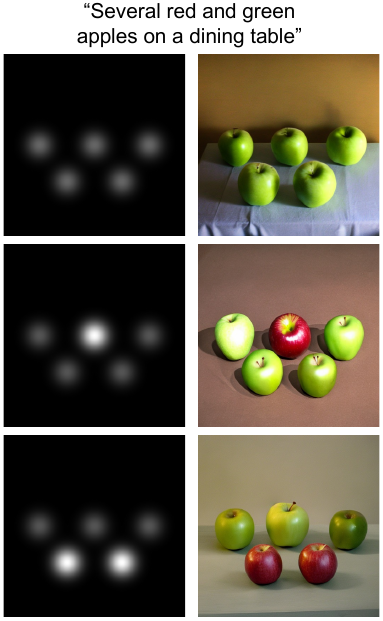}
  }
\Caption{Attention-controllable image generation with saliency guidance.}
{Given a text prompt and saliency map pair, \methodName generates images that not only contain the content as described by the text prompt but also attract viewers’ attention toward the desired image regions as emphasized by the saliency map. As illustrated in \subref{fig:teaser-all}, \revise{\methodName understands and exploits a diversity of factors inducing visual saliency, including low-level image features, such as color, contrast, frequency, orientation, and layout, and high-level semantic information, such as objects, texts, and faces;} \subref{fig:teaser-apple} demonstrates that \revise{\methodName can flexibly manipulate viewers' visual attention within the generated images by purely adjusting the color and contrast of specific image content while precisely following the user-specified layout}.}
\label{fig:teaser}
\end{figure}

Popular approaches to controlled image~\cite{zhang2023adding} and video~\cite{guo2023animatediff} generation include lightweight adaptation modules built around a foundation model. The adapter networks are usually conditioned by depth maps, semantic segmentation masks, body poses, or bounding boxes \cite{li2023gligen,zhang2023adding,ye2023ip,zhao2023uni} to control the spatial layout of an image or video. Meanwhile, an essential design factor for context creation is to direct viewers' attention to the regions of interest, such as buttons on web pages \cite{pang2016directing}, products being advertised \cite{bakar2015attributes}, or the storytelling events in a film \cite{shimamura2015attention}. Unlike the layout of an image, human attention is selective in nature~\cite{ungerleider2000mechanisms,kummerer2015information}, concurrently influenced by high-level semantics, mid-level layouts, and low-level visual features \cite{itti1998model,harel2006graph,kummerer2014deep,jia2020eml} (see examples in \Cref{fig:teaser}), as well as spatial and temporal characteristics \cite{droste2020unified,wang2018revisiting}. Therefore, existing conditioning mechanisms do not adequately control a viewer's visual spatial attention.

Our work aims to generate images and videos that guide viewers' visual attention toward specific regions of interest. To this end, we first analyze the spatial visual saliency~\cite{jia2020eml} of a large-scale image dataset \cite{jiang2015salicon}. The resulting image-saliency pairs are then leveraged to train a custom adapter network conditioned on saliency maps for text-to-image diffusion models. We further extend our saliency-aware conditioning from static images to temporally consistent video generation.

To evaluate the model's effectiveness in guiding real users' visual attention, we conducted a user study with human observers naturally examining the generated images with 3,000 eye-tracked trials. A series of objective evaluations demonstrate that our method consistently outperforms alternative control mechanisms in guiding user attention when viewing its generated visual content. We also showcase how the model can be applied in practical applications, including interactive designing saliency guidance, suppressing viewer attention in unwanted regions, and adapting generated content to various display and viewing conditions. This research serves as an important step toward human-centric control over generative models, focusing on the integration of typically implicit human design intentions within the content produced by these models. \revise{Our source code and pre-trained models are released at \url{https://github.com/NYU-ICL/saliency-guided-image-generation}.}

To summarize, our main contributions include:
\begin{itemize}
    \item introduce a saliency-guided image and video generation framework that incorporates the data priors of human visual attention mechanisms into the generation process;
    \item \revise{systematically study and validate the attention-directing performance of the proposed framework} through an eye-tracked user study and a large-scale model-based saliency analysis;
    \item demonstrate the availability and flexibility of the proposed framework in practical applications, including interactive design of saliency guidance, viewer attention suppression in unwanted regions, and adaptive visual content generation for varied display and viewing conditions.
\end{itemize}
\section{Related Work}
\label{sec:prior}


\subsection{Controllable Diffusion Models}

Recent advancements in diffusion-based generative models have found great success in image~\cite{rombach2022high} and video \cite{blattmann2023stable} content generation, as surveyed by Po et al.~\shortcite{po2023state} and Yang et al.~\shortcite{yang2023diffusion}. These foundation models, however, demand non-trivial prompt engineering to achieve desired user control over the generation process. To overcome this limitation, model customization approaches~\cite{hu2021lora,ye2023ip} as well as lightweight adapter networks~\cite{li2023gligen,zhang2023adding,zhao2023uni} have been established as the primary mechanisms for adding control over of the generated content. Current control strategies, however, use image-space annotations, including body pose, depth maps, or bounding boxes, to guide the \emph{spatial layout} of a generated image. \revise{While visual attention is impacted by spatial layout arrangements, it also largely depends on the interplay among low-level visual features (e.g., contrast, frequency, and color) and high-level semantic information (e.g., objects, texts, and faces).
Therefore, spatial attention exhibits selective and sometimes individually inconsistent patterns~\cite{kummerer2015information}. We use model-predicted saliency maps to condition a customized conditional diffusion model. This model is then used to steer the generation of images and videos to align with specific attention patterns based on the design intentions.}


\subsection{Human Visual Attention Modeling and Prediction}

Due to the complexity of cognitive visual attention \cite{ungerleider2000mechanisms}, modeling the saliency while perceiving images or videos has been an open challenge. Researchers have attempted to develop saliency models in a bottom-up fashion from image space statistical features \cite{judd2009learning,itti1998model,itti2001computational,harel2006graph,bruce2005saliency,bruce2007attention}. However, these low-level features by themselves are insufficient to account for top-down influences, e.g., our familiarity with different objects \cite{elazary2008interesting}. To measure these compounded influences, large-scale eye-tracked studies have been conducted, attempting to establish a paired image--video dataset with human-exhibited gaze fixations. Examples include MIT1003 \cite{judd2009learning}, CAT2000 \cite{borji2015cat2000}, SALICON \cite{jiang2015salicon,huang2015salicon} for images, VR saliency for 360 videos \cite{sitzmann2018saliency}, and DHF1K for videos \cite{wang2019revisiting}. These large-scale datasets catalyzed various deep neural network based saliency metrics for RGB images (e.g., DeepGaze models \cite{kummerer2014deep,kummerer2017understanding,linardos2021deepgaze}, EMLNet \cite{jia2020eml}, SalGAN \cite{pan2017salgan}), RGB-D frames \cite{sun2021deep,zhang2021uncertainty,ren2015exploiting}, videos (\cite{wang2018revisiting,droste2020unified,jiang2018deepvs,min2019tased}), and panoramas (\cite{zhang2018saliency}).
Saliency models are further extended to predict temporal fixation durations \cite{fosco2020much} and scanpaths \cite{martin2022scangan360}.


\subsection{Attention-Aware Computer Graphics}

Leveraging the selective attention distributions predicted by computational models has facilitated practical applications in enhancing the end-user experience or improving system efficiency. For instance, an existing image may be automatically enhanced by identifying and removing distractions without losing the fidelity \cite{mcdonnell2009eye,jiang2021saliency,mejjati2020look,aberman2022deep,miangoleh2023realistic}. High salient regions may also be prioritized for image loading \cite{valliappan2020accelerating}. Beyond images, character animation may also be enhanced with identified salient body parts \cite{mcdonnell2009eye}. So far, saliency-based optimization has been used to guide broad editing tasks given existing content. This research aims to introduce the building block of interactively creating desired images and videos from only simple user interference and text prompts.
\section{Method}
\label{sec:method}


\subsection{Saliency-guided Diffusion Model for Image Generation}
\label{sec:method-image-generation}

Diffusion models define a Markov chain that iteratively adds Gaussian noise to samples from an empirical data distribution and gradually converts them into noisy samples from a standard Gaussian distribution. They then learn the reverse diffusion (or denoising) process to iteratively remove noise and generate new data samples from randomly sampled Gaussian noise. State-of-the-art image diffusion models are commonly trained on large-scale image datasets and are capable of synthesizing visually appealing and content-diverse images \cite{po2023state}.

\begin{wrapfigure}{L}{0.55\textwidth}
\begin{center}
\includegraphics[width=0.53\textwidth]{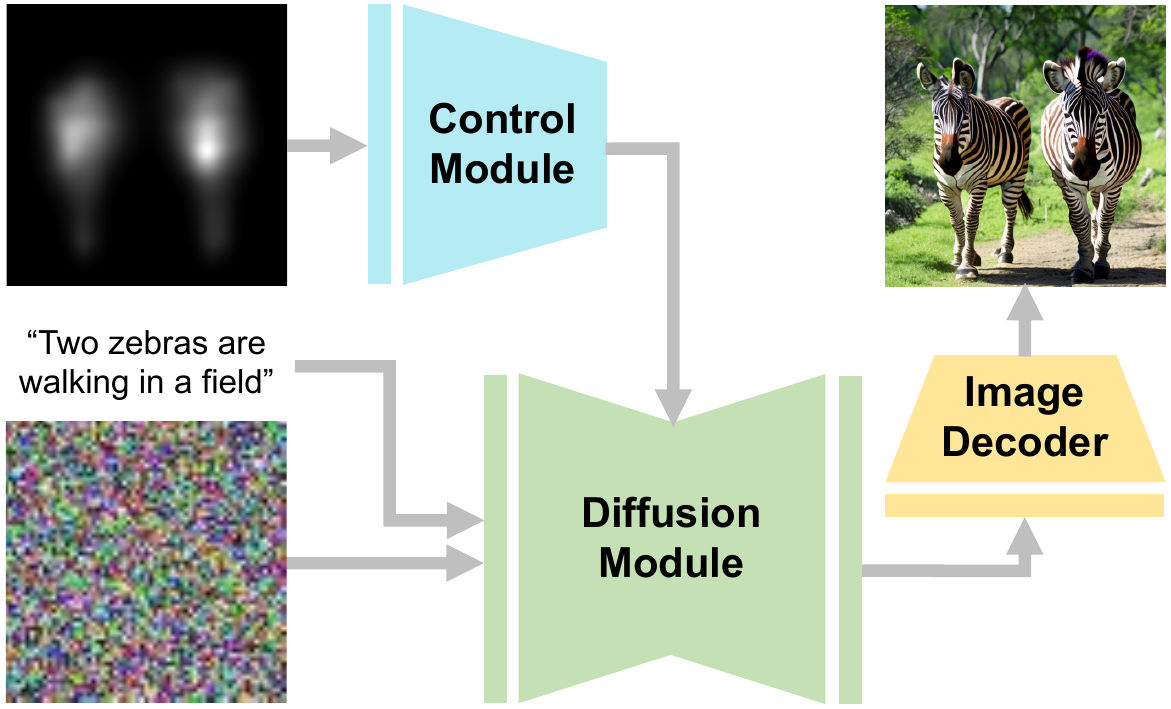}
\end{center}
\Caption{Saliency-guided image generation with \methodName.}
{Given randomly sampled noisy images as inputs, \methodName conditions the denoising process on user-specified saliency maps and text prompts such that the image features and semantic content in the generated images can trigger similarly distributed viewer attention.}
\label{fig:pipeline}
\end{wrapfigure}

Recent efforts toward controlling these large models, such as ControlNet \cite{zhang2023adding} and GLIGEN \cite{li2023gligen}, have shown that it is possible to incorporate a variety of multimodal conditions into the generation process. This integration allows for the manipulation of semantic information and spatial layout in the content produced by these large models. Such conditional generation is typically achieved by first augmenting pre-trained image diffusion models with an adaptation module and then fine-tuning on a considerably smaller set of condition-image pairs.

\begin{figure}[t]
\centering
\includegraphics[width=0.913\textwidth]{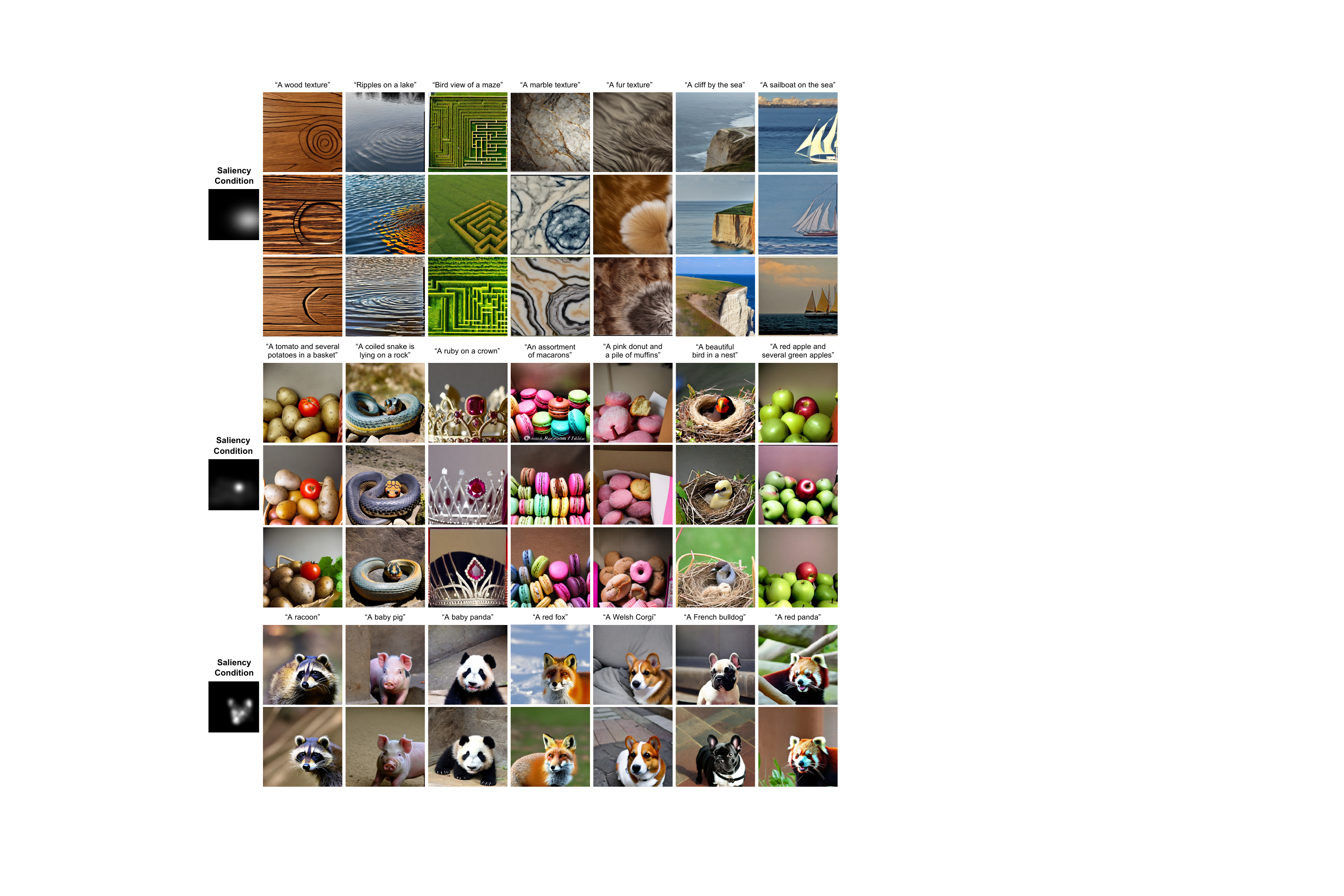}
\Caption{Saliency-guided image generation with \methodName.}
{\revise{During the generation process, \methodName leverages a variety of factors that affect visual saliency, such as color (e.g., the tomato and the apple scenes), frequency (e.g., the maze and the lake scenes), contrast (e.g., the fur and the marble texture scenes), orientation (e.g., the wood texture and the macaron scenes), layout (e.g., the last 3 rows), and high-level semantic information (e.g., the snake and the bird scenes).}}
\label{fig:evaluation-model-image-results}
\end{figure}

To achieve attention-aware image generation, we first curate a dataset of saliency-image pairs.
The scale of existing image datasets with paired eye-tracked human saliency data is commonly too limited ($<10k$ images) to support generative learning. Therefore, we leverage the captioned MSCOCO dataset \cite{lin2014microsoft} for images and a learning-based saliency model, EMLNet \cite{jia2020eml}, to predict their corresponding saliency maps. As visualized in \Cref{fig:pipeline}, we attach a ControlNet module to the encoder and middle blocks of a pre-trained Stable Diffusion (SD2.1) model \cite{rombach2022high} to inject saliency map conditions through zero convolutions and perform saliency-to-image translation. Particularly, the train, test, and unlabeled splits of MSCOCO 2017 (a total of 282k images) were taken to construct our training set $\mathcal{D}$. The 5k validation images of MSCOCO 2017 were held out for evaluation. \methodName was initialized with the pre-trained SD2.1 model checkpoint and finetuned using an Adam optimizer (constant learning rate $1e^{-5}$, $\beta_{1}=0.9$, and $\beta_{2}=0.999$) \cite{kingma2015adam} for $5e^{5}$ steps. Mathematically, our saliency-conditioned training process can be summarized as optimizing the denoising network in SD2.1, denoted as $\epsilon_{\theta}$, to predict the Gaussian noise added at each time step. \revise{The loss function driving the optimization can be formulated as follows:}
\begin{equation}
\mathcal{L}=\mathbb{E}_{z,t,c_{t},c_{s},\epsilon\sim\mathcal{N}(0,1)}\|\epsilon_{\theta}(z_{t},t,c_{t},c_{s})-\epsilon\|_{2}^{2}.
\end{equation}
where $z$, $z_{t}$, $c_{t}$, and $c_{s}$ denote a sample from the latent image distribution, the corresponding noisy sample after adding Gaussian noise for $t$ steps, the text prompt, and the conditioning saliency map, respectively.
\revise{As sampled cases in \Cref{fig:teaser,fig:evaluation-model-image-results} (and \Cref{fig:ablation-model-image-results,fig:ablation-saliency-changes} in the appendix) evidence, \methodName automatically incorporates a range of factors that lead to visual saliency, such as low-level image features (e.g. frequency, color, contrast, orientation, and layout) and high-level semantic information (e.g. objects, texts, and faces).}


\subsection{\revise{Extension to Saliency-guided Video Generation}}
\label{sec:method-video-generation}

\begin{figure}[t]
\centering
\includegraphics[width=0.99\textwidth]{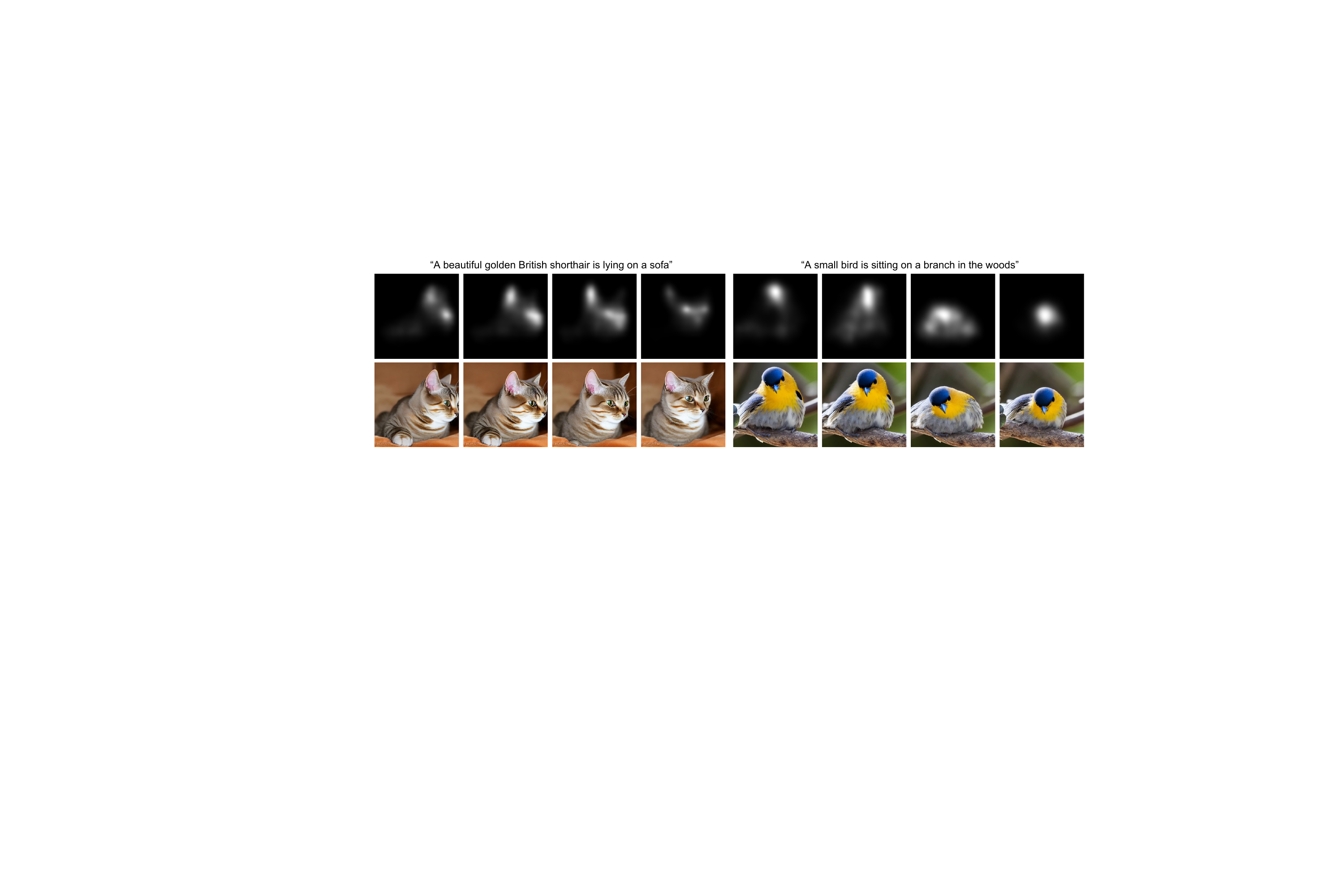}
\Caption{Saliency-guided video generation with \methodName.}
{\revise{By leveraging an off-the-shelf zero-shot video generation pipeline, \methodName can be extended to generate temporally consistent video clips with spatial-temporal saliency guidance.}}
\label{fig:evaluation-model-video-results}
\end{figure}

\revise{We further explore the possibility of applying \methodName to saliency-guided video generation. 
Existing zero-shot (i.e. training-free) video generation pipelines extend the generation of individual frames to video clips by enforcing the temporal consistency across adjacent frames} \cite{khachatryan2023text2video,guo2023animatediff}. 
Therefore, they are also adaptable to controlling modules with conditions such as body poses and edges. 

However, a unique challenge for human perception is the domain gap between viewing individual frames with extended duration (spatial-only saliency, $\saliencySpatialSeq=\saliencySpatial_{\{1,2,...,\timeDuration\}}$, as used in \Cref{sec:method-image-generation}) vs. watching them temporally composed as a video sequence (spatio-temporal saliency $\saliencyVideoSeq=\saliencyVideo_{\{1,2,...,\timeDuration\}}$) \cite{droste2020unified}. Here, $\saliencySpatial$ and $\saliencyVideo$ indicate the saliency maps on individual frames and videos of the same image sequence. This is due to various temporally induced factors -- such as camera and object motions -- influencing selective attention. Therefore, although it is possible to directly apply our control module to perform per-frame saliency-guided image generation $\saliencySpatialSeq$ to compose a temporally consistent video, the resulting user attention will differ from the target $\saliencyVideoSeq$. Concerning the domain gap, we adopt a video saliency prediction model, TASED-Net \cite{min2019tased}, to approximate $\saliencyVideoSeq$ for temporally consistent saliency sequences. \revise{\Cref{fig:evaluation-model-video-results} shows two sequences of \methodName-generated video frames.}
\section{Evaluation}
\label{sec:evaluation}

To quantitatively evaluate our saliency-guided approach to visual content generation, we conducted 1) an eye-tracked user study on \methodName-generated images to analyze its attention-directing performance (\Cref{sec:evaluation-user-study}); 2) a large-scale objective evaluation on \methodName-generated images and videos using off-the-shelf saliency models (\Cref{sec:evaluation-model}); 3) an ablation study on the quality and diversity of \methodName-generated images. 


\subsection{User Study: Tracking and Analyzing Viewers' Eye Gaze over Generated Images}
\label{sec:evaluation-user-study}

The aim of our research is to generate visual content that directs viewer attention in specific ways. This is achieved by incorporating the data priors of human visual attention into the generation process. To systematically analyze the attention-directing properties of the generated images, we conducted a user study with eye trackers to record participants' eye gaze patterns while they browse through a sequence of generated image samples.

\paragraph{Participants}
Twenty adults participated in the study (ages 23--57, 9 female). All of them have normal or corrected-to-normal vision, no history of visual deficiency, and no color blindness. None of them were aware of the hypothesis, the research, or the number of conditions. The research protocol was approved by the Institutional Review Board (IRB) at the host institution, and all subjects gave informed consent prior to the study.

\begin{wrapfigure}{L}{0.54\textwidth}
\begin{center}
\includegraphics[width=0.52\textwidth]{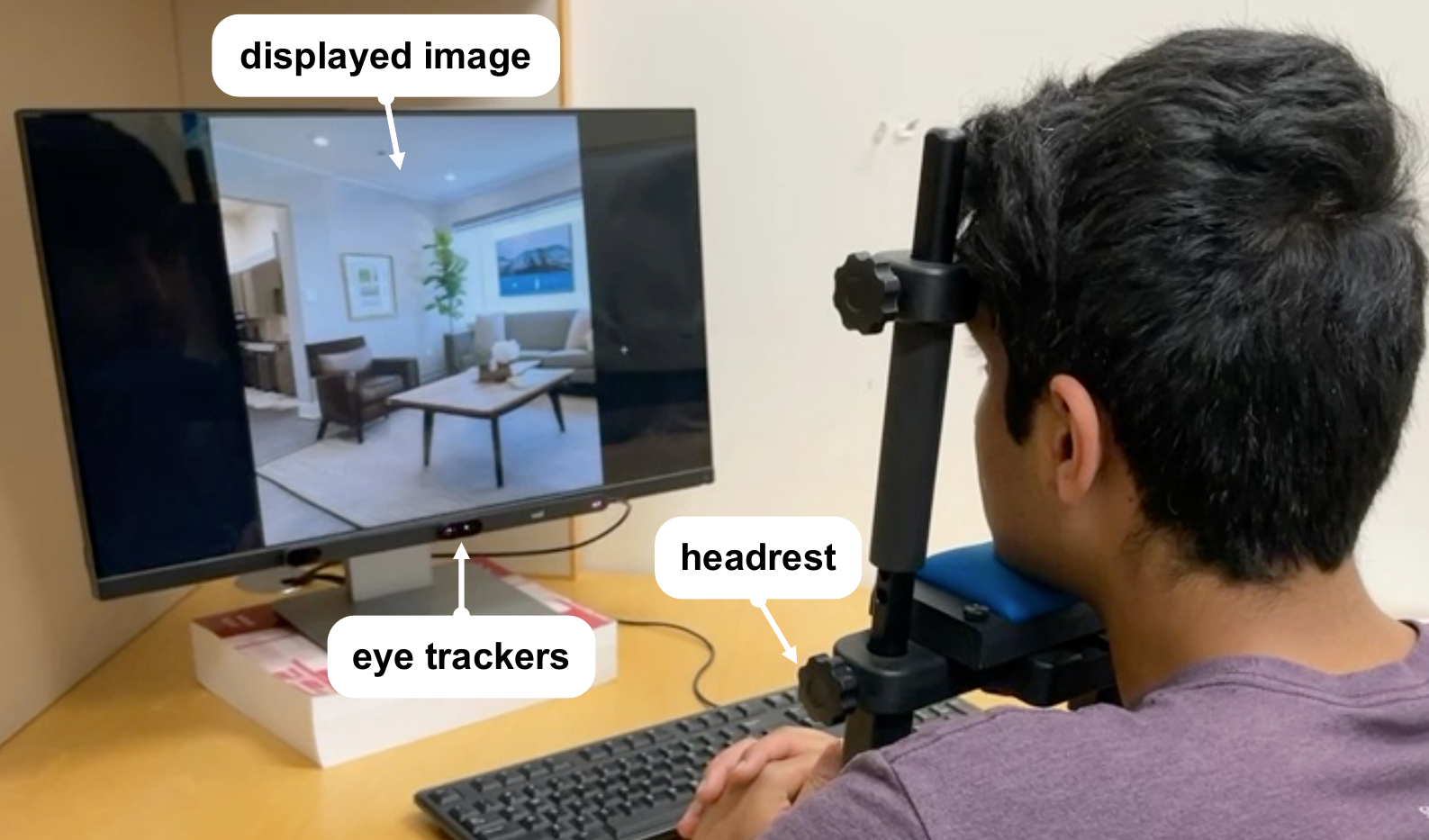}
\end{center}
\Caption{User study setup.}
{\revise{We used a Tobii Pro Spark eye tracker to record study participants' eye-gaze directions while they browsed through a sequence of generated images.}}
\label{fig:evaluation-user-study-setup}
\end{wrapfigure}

\paragraph{Setup and procedures}
During the study, subjects remained seated in a well-lit room and viewed a 24-inch Dell monitor (Model No. S2415H, resolution $1920\times1080$, luminance 250 $\text{cd}/\text{m}^{2}$) binocularly from an SR Research headrest positioned 60 cm away. The effective field of view and resolution were $46^{\circ} \times 26.8^{\circ}$ and 40 pixels per degree of visual angle. A Tobii Pro Spark eye tracker was mounted to the bottom of the monitor to record their eye gaze at 60 FPS. A 5-point eye-tracking calibration was performed before each session began. \Cref{fig:evaluation-user-study-setup} shows the experimental setup of our user study.

\paragraph{Stimuli}
We first sampled 50 images from the held-out validation set of MSCOCO 2017, where half of them have humans/animals as the main content and the rest show close-up shots of objects or nature/city scenes. These selected images were annotated using the BLIP-2 Image2Text model \cite{li2023blip} for text prompt conditioning. Image saliency maps were extracted using the EML-Net saliency model \cite{jia2020eml} for visual saliency conditioning. We then fed the obtained paired text prompts and saliency maps to our saliency-guided model to generate 50 images of $512\times512$ resolution as the visual stimuli for the user study. The hypothesis is that these generated images should direct viewers’ attention toward the intended regions as depicted by the saliency maps while observing the text prompts and maintaining non-degraded image quality/diversity.

\paragraph{Conditions}
\revise{For attention-directing performance comparison, we included two baseline conditions, the text-only model ($\conditionUNCOND$), Stable Diffusion v2.1 \cite{rombach2022high}, and the bounding-box-guided model ($\conditionBBOX$), GLIGEN \cite{li2023gligen}.} All three conditions share the same input text prompts to control what visual content should be generated. Additionally, $\conditionBBOX$ takes in text-annotated bounding boxes predicted by the Grounding DINO model \cite{liu2023grounding}, and $\conditionSMAP$ (i.e. \methodName) takes in saliency maps predicted by the EML-Net saliency model. Similar to $\conditionSMAP$, 50 images were generated for $\conditionUNCOND$ and $\conditionBBOX$, respectively.

\begin{figure}[t]
\centering
\includegraphics[width=0.95\textwidth]{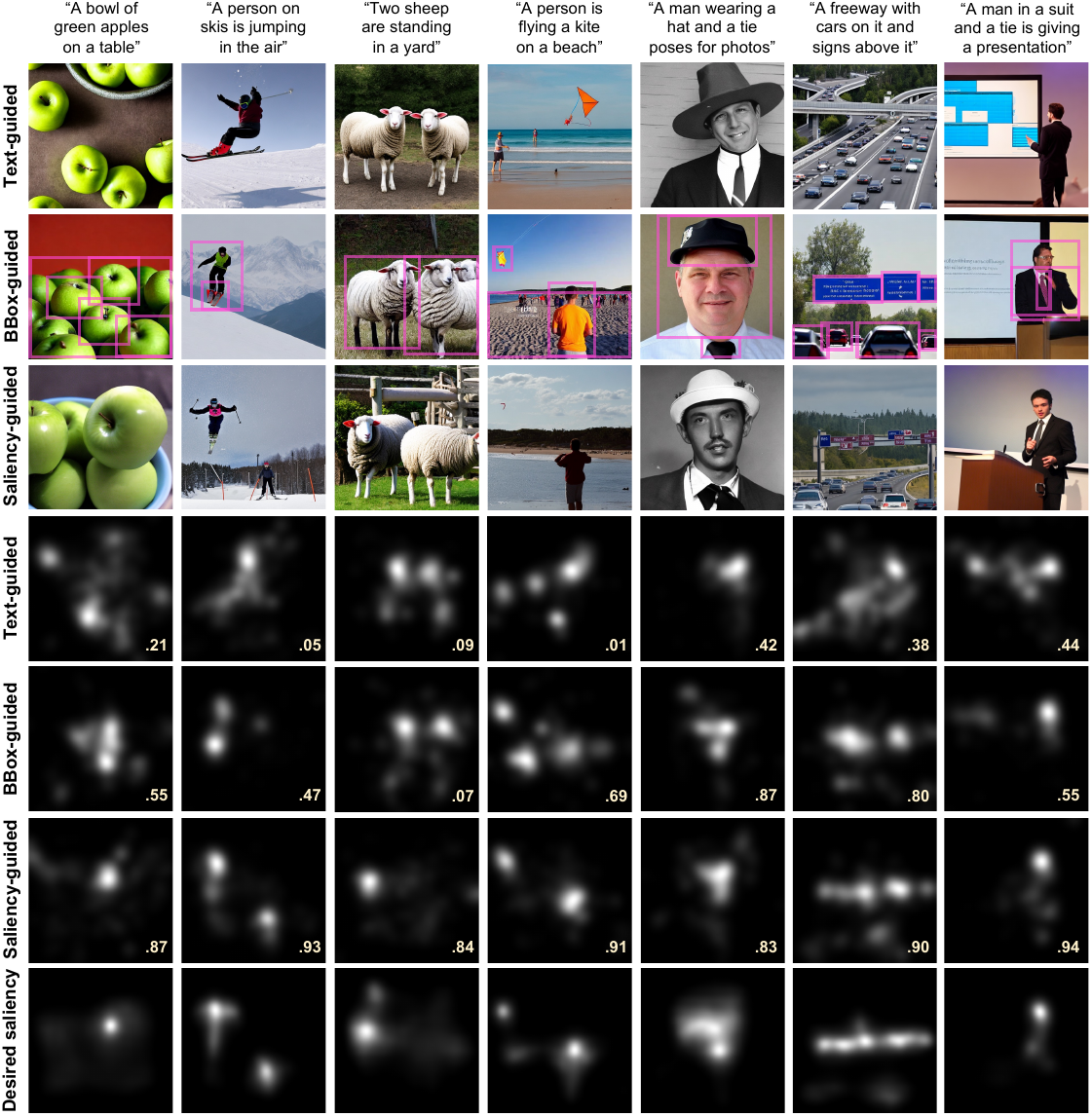}
\Caption{Eye-tracked user study.}
{Rows 1--3 show the generated images, rows 4--6 show the empirical saliency maps obtained by aggregating 20 users' eye gaze data, and row 7 shows the input conditioning saliency maps (i.e., the desired viewer attention distributions). The conditioning bounding boxes for \conditionBBOX are shown as pink overlays. The number associated with each empirical saliency map shows its correlation with the desired attention distribution. Compared to the two baseline methods, \revise{the images generated by \methodName ($\conditionSMAP$) not only contain the exact content as described by the text prompts but also trigger viewer attention distributions that align better with the intended ones.}}
\label{fig:evaluation-user-study-results}
\end{figure}

\paragraph{Task and duration}
The total of 150 images generated for the three conditions was shuffled in random order and sequentially displayed to each subject, with a 5-second duration for each image and a 1-second pause between consecutive images. The complete study, including hardware setup/calibration, pre-study instructions, and breaks, took about 30 minutes per subject. Throughout the study, all subjects were instructed to keep their head stationary on the headrest and freely explore the displayed images by shifting their eye gaze. Their eye gaze patterns on each image were recorded to compute the corresponding empirical saliency map.

\paragraph{Metrics}
To quantitatively evaluate each model's performance in directing users' attention to intended image regions, we adopted five saliency similarity metrics from the MIT/Tuebingen Saliency Benchmark \cite{kummerer2018saliency}: Area Under ROC Curve (AUC) \cite{kummerer2015information}, Normalized Scanpath Saliency (NSS) \cite{peters2005components}, Correlation Coefficient (CC), Kullback–Leibler Divergence (KL), and Histogram Intersection (SIM). Notably, AUC and NSS take a saliency map and a sequence of eye fixations as inputs, while computing CC, KL, and SIM requires two saliency maps. To convert our collected eye-tracking data into empirical saliency maps, we followed the same post-processing procedures described in \cite{sitzmann2018saliency}.

\begin{table}[t]
\centering
\Caption{Eye-tracked user study.}
{\revise{\methodName outperforms the two baseline methods in directing viewers' visual attention toward the user-intended image regions.} $\uparrow$/$\downarrow$ indicates that higher/lower score is better.}
 \begin{tabular}{c | c c c c c}
 \hline
  & $\text{AUC}\uparrow$ & $\text{NSS}\uparrow$ & $\text{CC}\uparrow$ & $\text{KL}\downarrow$ & $\text{SIM}\uparrow$ \\
 \hline
 $\conditionUNCOND$ & 0.65 & 0.71 & 0.21 & 4.75 & 0.34 \\
 $\conditionBBOX$ & 0.78 & 1.21 & 0.47 & 2.67 & 0.48 \\
 $\conditionSMAP$ & \textbf{0.84} & \textbf{1.82} & \textbf{0.78} & \textbf{0.79} & \textbf{0.66} \\
 \hline
 \end{tabular}
\label{tab:evaluation-user-study}
\end{table}

\paragraph{Results and discussion}
As shown in \Cref{tab:evaluation-user-study}, our saliency-guided approach consistently outperforms the two baselines in controlling and directing users' visual attention toward intended image regions across all 5 metrics. \Cref{fig:evaluation-user-study-results} shows seven groups of generated images used in our user study, their corresponding text prompts and empirical saliency maps, as well as the input conditioning saliency maps. As can be observed, \methodName not only produces the content as depicted by the text prompts but also achieves viewer attention distributions that align well with the intended ones. These results strongly validate the attention-directing capability of \methodName.


\subsection{Model-based Saliency Analysis}
\label{sec:evaluation-model}

In addition to the eye-tracked user study, we further performed a large-scale objective analysis of \methodName-generated images and videos to more comprehensively understand its capabilities. The two previously introduced baseline methods, \conditionUNCOND and \conditionBBOX, are again adopted for comparisons.

\paragraph{Image}
Similar to the data preparation procedures in \Cref{sec:evaluation-user-study}, we first computed the BLIP-2 captions, spatial saliency maps, and text-annotated bounding boxes on the held-out MSCOCO data (5K images) to condition the image generation process. Next, we applied \methodName to generate 5K images for \conditionSMAP using the 5K paired captions and saliency maps. Using their respective model and input conditions, 5K images were also generated for \conditionBBOX and \conditionUNCOND. Finally, we measured the discrepancy between the saliency maps used as input conditions (i.e., the desired attention distributions) and the saliency maps of the generated images per the EMG-Net image saliency predictor (i.e., the achieved attention distributions).
 
\paragraph{Video}
To generate a large set of video clips with \methodName, we first sampled 1K videos from the WebVid-10M dataset \cite{bain2021frozen}, trimmed them down to 4-second clips, and uniformized the frame rate to 8 FPS. Next, we took the TASED-Net video saliency predictor to extract the spatial-temporal saliency map sequence from each processed video clip. Leveraging the Text2Video-Zero pipeline \cite{khachatryan2023text2video}, we then applied our \methodName model to generate 1K video clips of 4 seconds and 8 FPS based on the previously extracted saliency map sequences. Finally, similar to generated images, we measured the frame-wise discrepancy between the desired and achieved attention distributions for all generated video clips.

\paragraph{Results and discussion}
In this large-scale evaluation, since we take advantage of ML-based saliency models to directly extract saliency maps from generated images/videos as ground truth and do not have access to eye fixation data, only CC, KL, and SIM are feasible and thus adopted. As shown in \Cref{tab:evaluation-model}, \methodName significantly outperformed the two baselines across all three metrics on both image and video generation. The analysis above not only validates the extendability of \methodName to saliency-guided video generation but also demonstrates its robustness and generality on a wide range of text prompts and attention distributions. These results laid the foundations for the practical applications that we discuss in \Cref{sec:applications}.

\begin{table}[t]
\centering
\Caption{Model-based saliency analysis.}
{The images and videos generated by \methodName achieve more aligned visual attention distributions with the input saliency maps according to EML-Net and TASED-Net. Note that the video results of \conditionBBOX are unavailable due to GLIGEN's incompatibility with the Text2Video-Zero pipeline. 
\nothing{The subscripts $I$ and $V$ indicate the results for generated images and videos, respectively.}}
 \begin{tabular}{c | c c c | c c c}
 \hline
  & & Image & & & Video & \\ 
 \hline
  & $\text{CC}\uparrow$ & $\text{KL}\downarrow$ & $\text{SIM}\uparrow$ & $\text{CC}\uparrow$ & $\text{KL}\downarrow$ & $\text{SIM}\uparrow$ \\
 \hline
 \conditionUNCOND & 0.22 & 7.27 & 0.35 & 0.21 & 5.68 & 0.34 \\
 \conditionBBOX & 0.54 & 3.97 & 0.54 & N/A & N/A & N/A \\
 \conditionSMAP & \textbf{0.84} & \textbf{0.75} & \textbf{0.75} & \textbf{0.79} & \textbf{0.85} & \textbf{0.68} \\
 \hline
 \end{tabular}
 \label{tab:evaluation-model}
\end{table}


\subsection{Ablation Study on Image Quality and Diversity}
\label{sec:evaluation-ablation}

To verify that our saliency conditioning does not result in any performance drop to the base SD2.1 model, we adopted two image quality/diversity metrics, inception score (IS, higher is better) and Fréchet inception distance (FID, lower is better), to evaluate how do \methodName-generated images compare with those generated by the base model. To this end, we reused the two sets of images generated for the conditions \conditionUNCOND (the base model with text guidance only) and \conditionSMAP (our \methodName model). Note that both sets of images were generated using the same text prompts and are thus fair to be compared. Notably, \methodName achieved $\text{FID}=16.43$ and $\text{IS}=36.05$, while SD2.1 achieved $\text{FID}=22.29$ and $\text{IS}=34.81$, indicating that our saliency-guided generation even improves upon the base model in terms of image quality and diversity. Such improvements are likely due to the similar image distribution of our training and evaluation sets, which were both taken from MSCOCO 2017. Our \methodName model was carefully fine-tuned on these photo-based realistic images to learn saliency guidance while SD2.1 was not specifically optimized to generate such images. These results strongly validate that we can effectively incorporate the data priors of human attentional behaviors into the diffusion process and generate saliency-guided images while not degrading the image generation performance of the base diffusion models.
\section{Applications}
\label{sec:applications}


\subsection{\revise{Interactive Design of Saliency Guidance}}
\label{sec:applications-design}

\begin{figure}[t]
\centering
\includegraphics[width=0.85\textwidth]{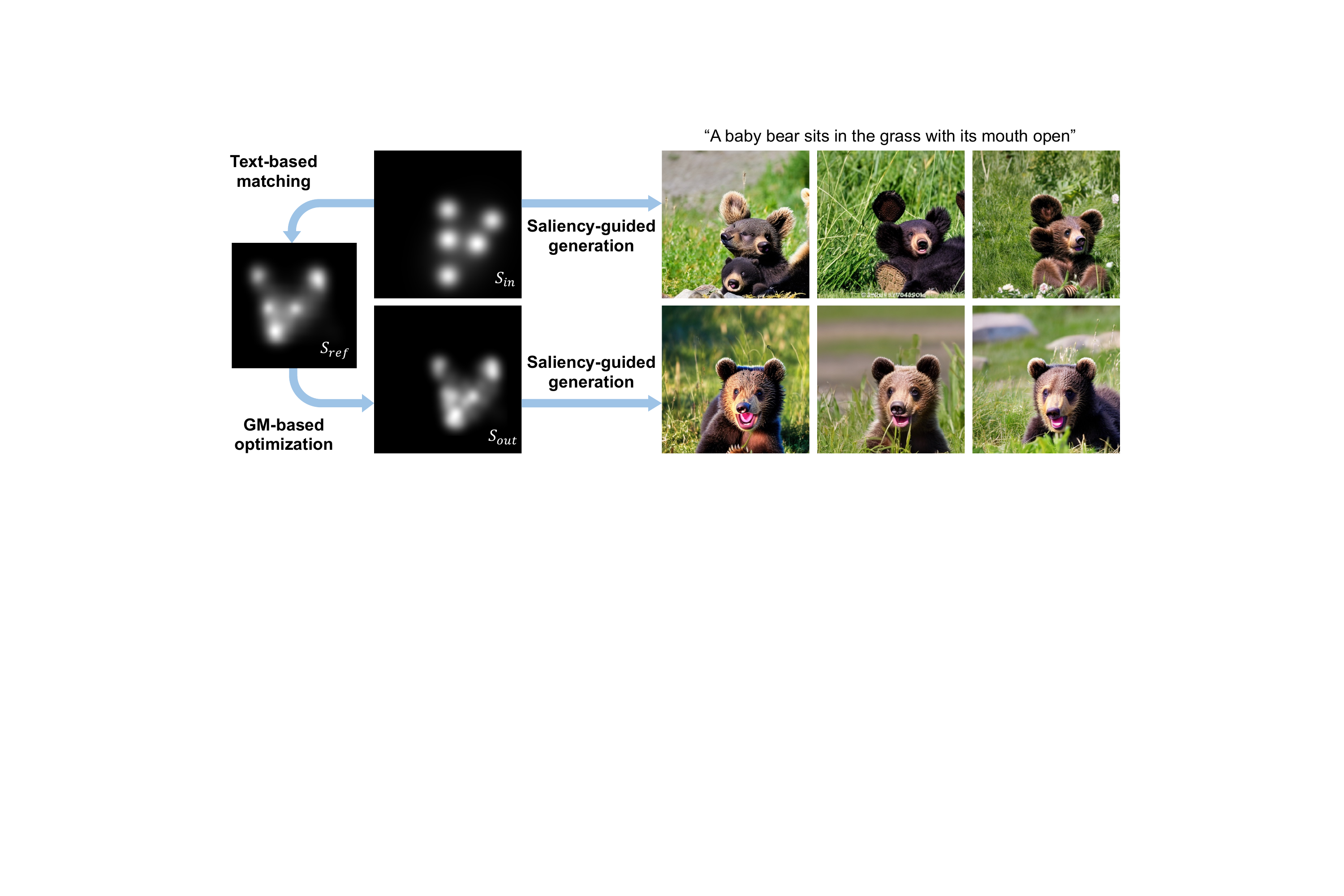}
\Caption{Interactive design of saliency guidance.}
{\revise{Our interactive saliency design system eases users' burden of crafting high-quality, prompt-compatible saliency maps by proposing corrections when necessary. The top row shows a tentative user-created saliency map and its corresponding \methodName-generated images; the bottom row shows the artifact-free generation results obtained after optimizing the saliency guidance.}}
\label{fig:applications-design}
\end{figure}

Unlike other non-semantic controlling inputs, such as body poses, line sketches, and depth maps,
prior research has observed a strong correlation between the content of an image and its incurred visual saliency \cite{borji2015salient}. 
Consequently, when the text prompt is incompatible with the saliency guidance provided by a designer, \methodName may generate unnaturally looking images with artifacts (e.g., distorted objects and human bodies, see the top row of \Cref{fig:applications-design}), as we have repeatedly observed in our experiments.

\revise{To this end, we developed an interactive saliency authoring system to ease designers' trial-and-error efforts in crafting prompt-compatible saliency guidance. Specifically, a user provides a text prompt $\textPrompt_{\text{in}}$ and generates a tentative saliency map by clicking over a black canvas. Each click creates a bivariate Gaussian $\gaussian_{i}(\gaussianWeight_i,\gaussianMean_i,\gaussianCov_i)$ that can be further adjusted to compose the desired saliency guidance.} Here, $\gaussianWeight_i$, $\gaussianMean_i$, and $\gaussianCov_i$ denote the Gaussian's weight, mean, and covariance matrix.
The resulting saliency map is represented as a Gaussian mixture (GM):
\begin{align}
\saliencySpatial_{\text{in}} \coloneqq \sum_{i=1}^{N_{\text{in}}}\gaussian_{i}\left(\gaussianWeight_i,\gaussianMean_i,\gaussianCov_i\right).
\end{align}
\revise{Next, from our 282k text-saliency paired dataset $\mathcal{D}$ (\Cref{sec:method-image-generation}), we search for the text prompt closest to the user-provided one within the CLIP-embedded language space \cite{radford2021learning}, and retrieve its corresponding saliency map (i.e. the reference saliency map) as a starting point for optimization:}
\begin{align}
\textPrompt_{\text{ref}}, \saliencySpatial_{\text{ref}} = \underset{\{\textPrompt,\saliencySpatial\}\in\mathcal{D}}{\arg\min}\, \biggl\|\,\mathcal{E}(\textPrompt)-\mathcal{E}(\textPrompt_{\text{in}})\,\biggr\|_{2}.
\end{align}
Here, $\textPrompt_{\text{ref}}$ and $\saliencySpatial_{\text{ref}}$ denote the retrieved reference text-saliency pair; $\mathcal{E}$ denotes the CLIP text encoder.
\revise{Finally, we convert $\saliencySpatial_{\text{ref}}$ into a Gaussian mixture $\hat{\saliencySpatial}_{\text{ref}}$ and optimize an image-space transformation matrix $T \in \mathbb{R}^{2\times2}$, which is composed of a translation vector $t \in \mathbb{R}^{2}$, a rotation angle $r \in [0, \pi]$, and a scaling vector $s \in \mathbb{R}_{+}^{2}$, to approximate the user-intended saliency guidance $\saliencySpatial_{\text{in}}$ as follows:
\begin{align}
& \hat{\saliencySpatial}_{\text{ref}} \coloneqq \sum_{j=1}^{N_{\text{out}}}\gaussian_{j}\left(\gaussianWeight_j,\gaussianMean_j,\gaussianCov_j\right), \\
\mathcal{L} = \left\lVert\,T \cdot \hat{\saliencySpatial}_{\text{ref}}-\saliencySpatial_{\text{in}}\,\right\rVert_{2}, & \quad
T_{\text{out}} = \underset{T \in \mathbb{R}^{2\times2}}{\arg\min} \, \mathcal{L}, \quad
\saliencySpatial_{\text{out}} = T \cdot \hat{\saliencySpatial}_{\text{ref}}.
\end{align}
Here, $\mathcal{L}$, $T_{\text{out}}$, and $\saliencySpatial_{\text{out}}$
denote the objective function, optimized transformation matrix, and corrected saliency map, respectively.} \Cref{fig:applications-design} shows an example where the initial user-specified saliency map $\saliencySpatial_{\text{in}}$ is incompatible with the accompanying text prompt due to inappropriate saliency intensity and layout. 
In the corresponding generated images, the baby bear exhibits a severely disfigured face and incorrect body anatomy. 
In comparison, the images generated based on the system-corrected saliency map $\saliencySpatial_{\text{out}}$ not only present a realistically looking baby bear without artifacts but also maintain the user-intended saliency layout. 
Notably, the whole generation process only involves a few mouse clicks from the users with the correction automatically performed. Using this interactive design interface, users can easily create complicated custom saliency maps to perform attention-controlled image generation (refer to \Cref{fig:teaser,fig:evaluation-model-image-results} for a few examples).


\subsection{Attention Suppression}
\begin{figure}[t]
\centering
\includegraphics[width=0.75\textwidth]{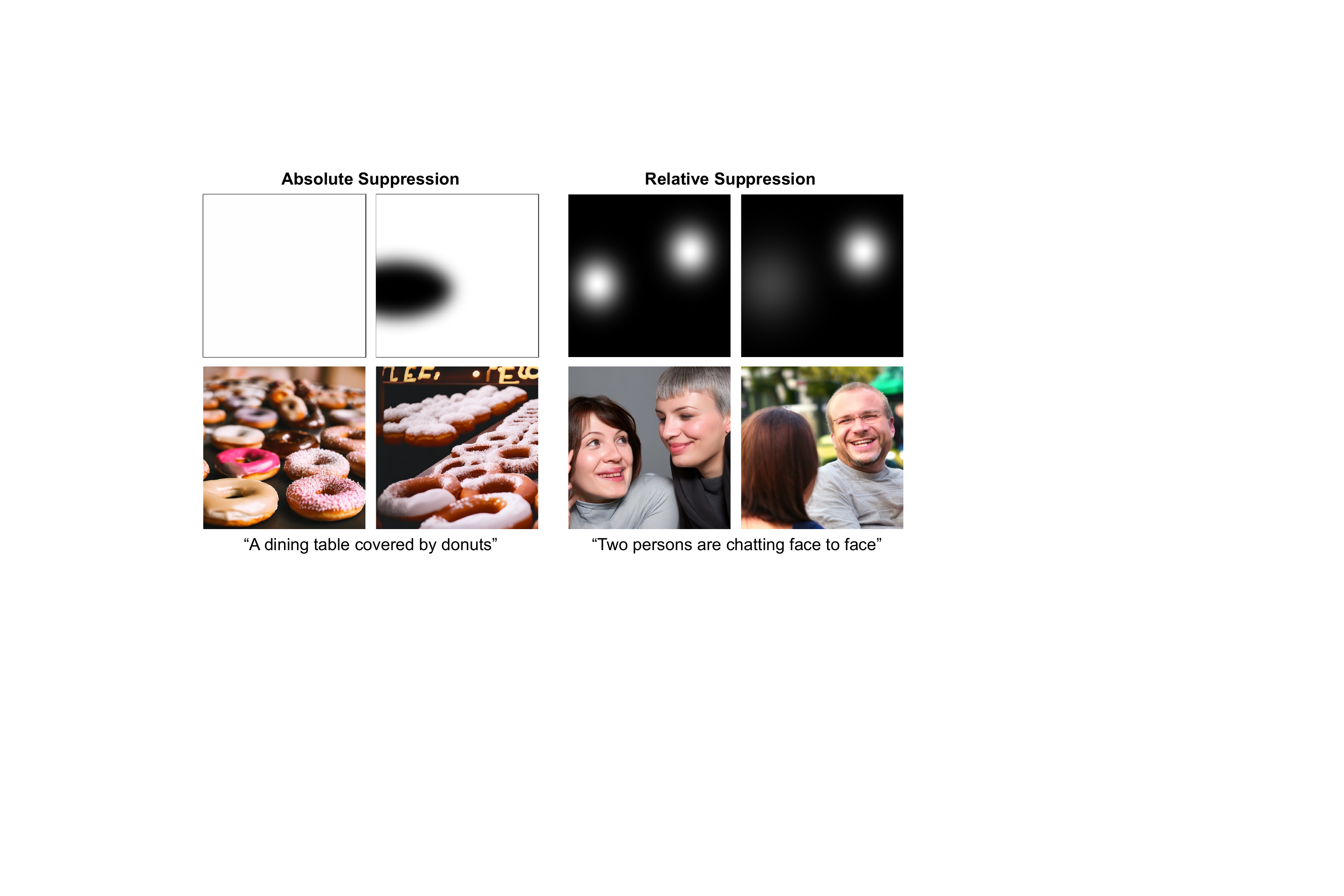}
\Caption{Demonstration of attention suppression.}
{\revise{\methodName is capable of completely shifting viewers' attention away from specific image regions (absolute suppression) or reducing viewers' attention in less important regions (relative suppression). Brighter/darker colors indicate higher/lower saliency (or desired attention levels).}}
\label{fig:applications-suppression}
\end{figure}
Now that we have demonstrated \methodName's capabilities in generating high-quality images and videos that attract viewers' attention toward the intended regions, we also explore the opposite effects of suppressing viewers' attention in unwanted regions. 
In particular, we differentiate between two types of attention suppression using \methodName: 
1) absolute suppression, where viewer attention is completely shifted away from the target regions; 
2) relative suppression, \revise{where viewer attention is adaptively reduced in the less important regions compared to the more important ones, though not entirely removed.}

As demonstrated in \Cref{fig:applications-suppression}, we crafted two pairs of saliency maps and fed them to \methodName. In each pair, one saliency map is designed to demonstrate a type of attention suppression (\groupTEST), and the other is crafted to be the same except in the suppressed regions (\groupCONTROL). For 1) \emph{absolute suppression}: \groupTEST is strongly salient everywhere except for an oval-shaped region. The \groupTEST-guided image features several arrays of donuts except for the suppressed region, while the \groupCONTROL-guided image shows donuts everywhere.
For 2) \emph{relative suppression}, the text prompt describes a two-person chatting scenario. \groupCONTROL shows two separate bright regions that are equally salient, and \groupTEST shows the same two bright regions but with one of them being considerably brighter than the other. While the \groupCONTROL-guided image depicts two persons at the two bright regions, the \groupTEST-guided image, interestingly, not only positions the two persons to the two bright regions but also makes the person at the brighter one face the camera and the other person turn his back to the camera and out of focus. \revise{More attention suppression results can be found in \Cref{fig:ablation-absolute-suppression,fig:ablation-relative-suppression} in the appendix.}


\subsection{Display-Adaptive Generation}
\begin{figure}[t]
\centering
\includegraphics[width=0.6\textwidth]{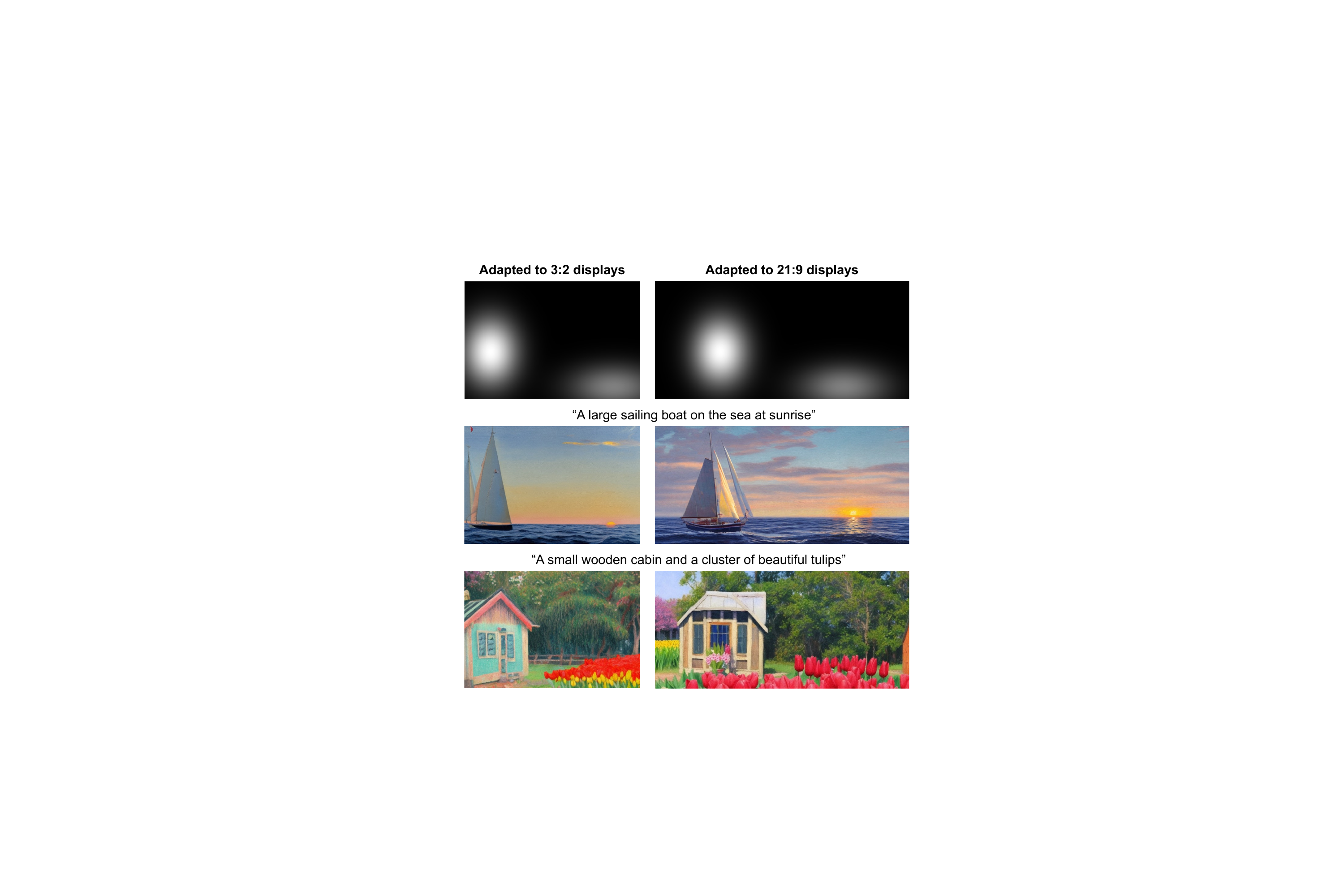}
\Caption{Demonstration of display-adaptive image generation.}
{\revise{The left/right column shows the target saliency distribution and the corresponding generated images for a narrow/wide field display. Notably, the spatial guidance for the wide-field display is designed to ensure that the main image content remains in a comfortable FoV for viewers. This can help improve content visibility and reduce excessive eye/head movements, countering the low visual acuity of peripheral vision.}}
\label{fig:applications-display}
\end{figure}
\revise{The increasing size of the latest displays has widened the field of view (FoV) they deliver. However, most visual content is created without considering the exact environment in which it will be displayed. In practice, this may lead to unpredictable and compromised viewing experiences. For instance, when viewing an image on a 13-inch MacBook, most of the image content will be located within the central part of the viewers' visual field; when viewing the same image on a 34-inch curved monitor with a 21:9 aspect ratio, however, a non-trivial part of the image will be displayed in the viewers' low-acuity peripheral visual field \cite{eriksen1985allocation,reeves1999effects} and not be properly consumed. In addition, viewers also have to shift their eye gaze or rotate their necks more frequently, which may lead to ergonomic problems \cite{gallagher2021does,zhang2023toward}.}

\revise{Leveraging \methodName, artists and designers can readily customize their desired saliency guidance and generate graphics that are perceptually optimized to varying displays (aspect ratio and size) and viewing conditions (viewing distance and angle). This feature can help, for instance, keep image content of particular interests and importance within a reasonable FoV for viewers to enjoy an uncompromised viewing experience when switching from one display to another. As demonstrated in \Cref{fig:applications-display}, \methodName-based adaptive generation shifts important image content to the optimal viewing angle for the display in use through saliency conditioning, effectively improving content visibility and reducing the necessity of frequent eye/head movements due to peripheral viewing.}
\section{Limitations and Future Work}

Our saliency-guided controlling module was established based on an end-to-end computational saliency model optimized on human gaze data \cite{jia2020eml}.
However, the factors inducing human visual attention are diverse and multi-dimensional, including low-level image features, mid-level local structures, and high-level semantics \cite{hayes2021deep}. 
Currently, \methodName model does not differentiate between these underlying causes.
Integrating various saliency models targeting different levels of saliency-triggering factors under a probabilistic framework \cite{kummerer2015information} may shed light on more fine-grained saliency-based control, such as specifying local color- and contrast-induced saliency vs. semantic labels in the saliency map.

The extension to saliency-guided video generation, while showing numerical saliency alignments between the input control and the computed prediction with TASED-Net \cite{min2019tased}, has not been measured with human observers. This is due to the significantly large sampling requirement to obtain an eye-tracking-revealed spatial-temporal saliency \cite{wang2018revisiting}. 
To this end, we plan to investigate eye-tracking-free human attention assessment approaches via crowdsourcing platforms, e.g., \cite{kim2017bubbleview}.
In addition, the current approach treats the saliency frames in the control sequence separately to generate temporally consistent videos frame-by-frame with \cite{khachatryan2023text2video}. 
Introducing an explicit temporal module to exploit the temporally induced factors, such as motions between adjacent frames, may further improve the controlling effectiveness.

The proposed interactive design tool automatically corrects users' arbitrarily specified saliency input through a text-saliency compatibility matching and adjustment, as shown in \Cref{fig:applications-design}.
Recognizing that image artifacts may also influence saliency \cite{yang2021measurement}, we plan to investigate the three-fold cross effects among characterized saliency maps, text feature spaces, and their generated image quality assessment \cite{golestaneh2022no,zhang2014vsi,yang2019sgdnet}. 
A quantifiable correlation in the continuous domain may provide guidance for quality-predictable and saliency-aware generation.

\section{Conclusion}
In this paper, we present a saliency-guided generative model that guides users' viewing attention. 
It may match designers' specifications with a simple click-and-run user interface. 
An eye-tracked user study evidences the real-world effectiveness. 
With various demonstrated applications, such as inversely suppressing attention and adapting the generations to various viewer-display conditions, we hope the research will initiate the first step of viewer-perception-aware generative models. 

\begin{acks}
This research is partially supported by the National Science Foundation grant \#2232817 and a gift from Google.
\end{acks}


\bibliographystyle{ACM-Reference-Format}
\bibliography{paper.bib}
\clearpage
\begin{figure}[h]
\centering
\includegraphics[width=0.9\textwidth]{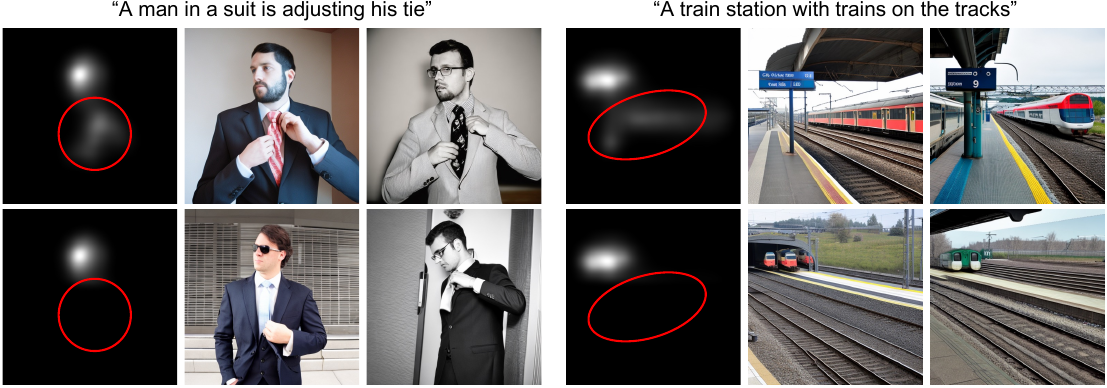}
\Caption{\revise{Ablation study on subtle changes in saliency guidance.}}
{\revise{Visually less significant components in the conditioning saliency map (highlighted regions) may have significant impacts on the generation results.}}
\label{fig:ablation-saliency-changes}
\end{figure}
\begin{figure}[h]
\centering
\includegraphics[width=0.85\textwidth]{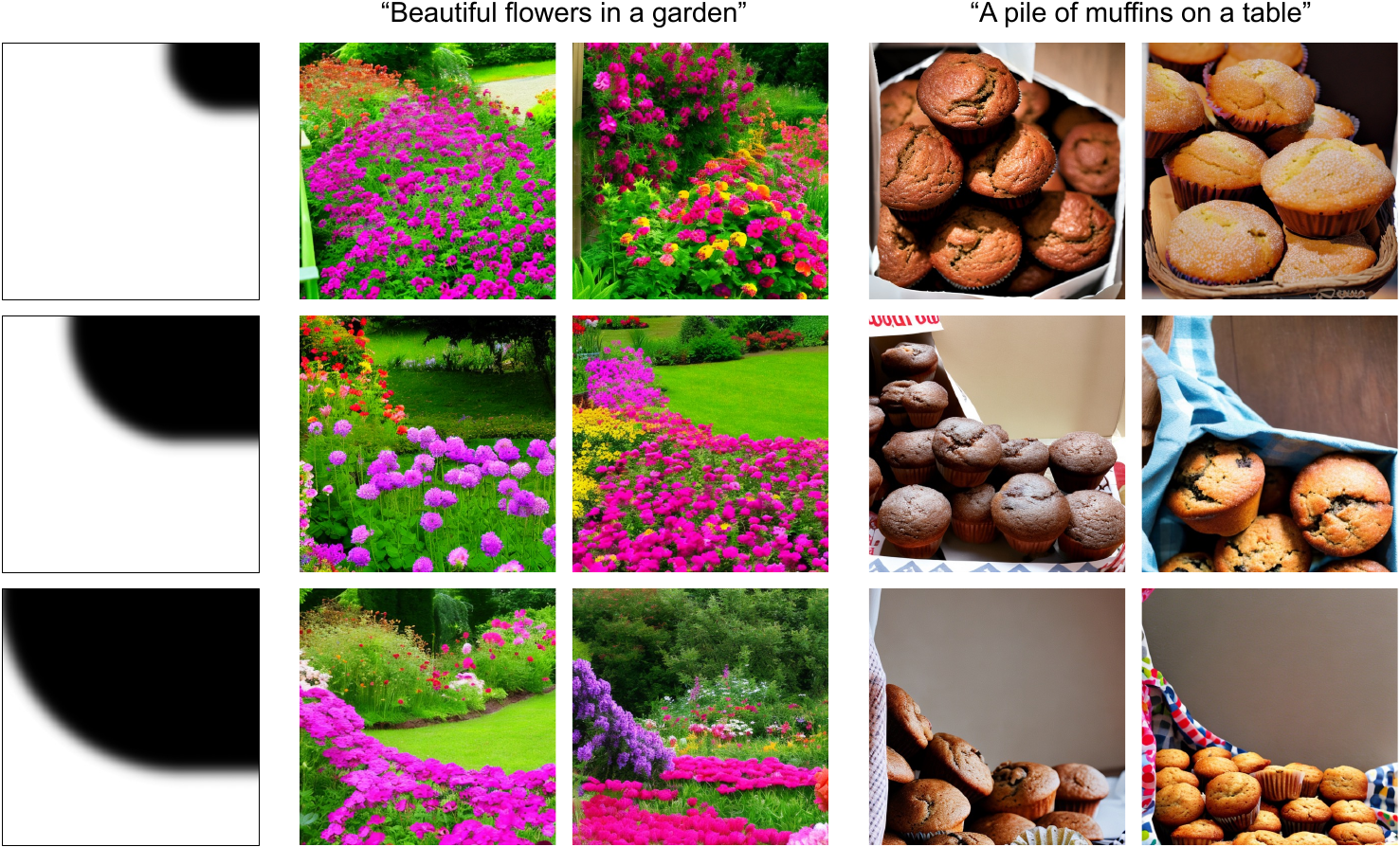}
\Caption{\revise{Absolute attention suppression with \methodName.}}
{}
\label{fig:ablation-absolute-suppression}
\end{figure}
\begin{figure}[h]
\centering
\includegraphics[width=0.85\textwidth]{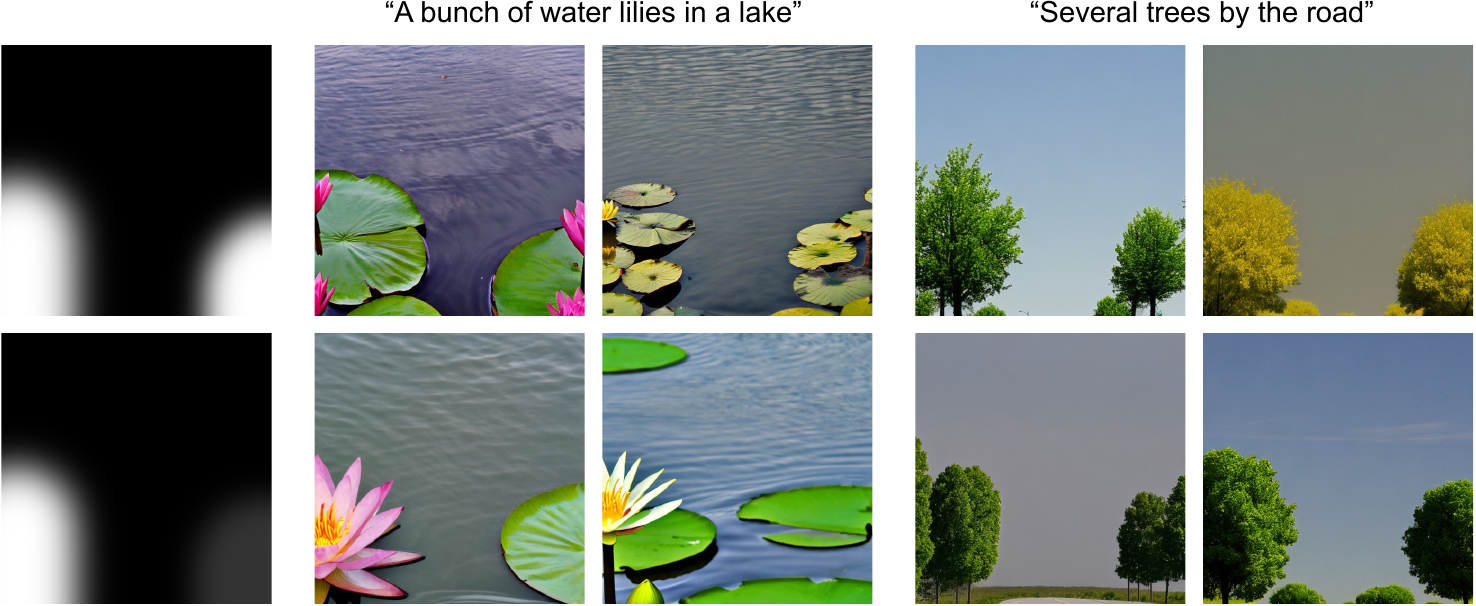}
\Caption{\revise{Relative attention suppression with \methodName.}}
{}
\label{fig:ablation-relative-suppression}
\end{figure}

\begin{figure}[h]
\centering
\includegraphics[width=0.9\textwidth]{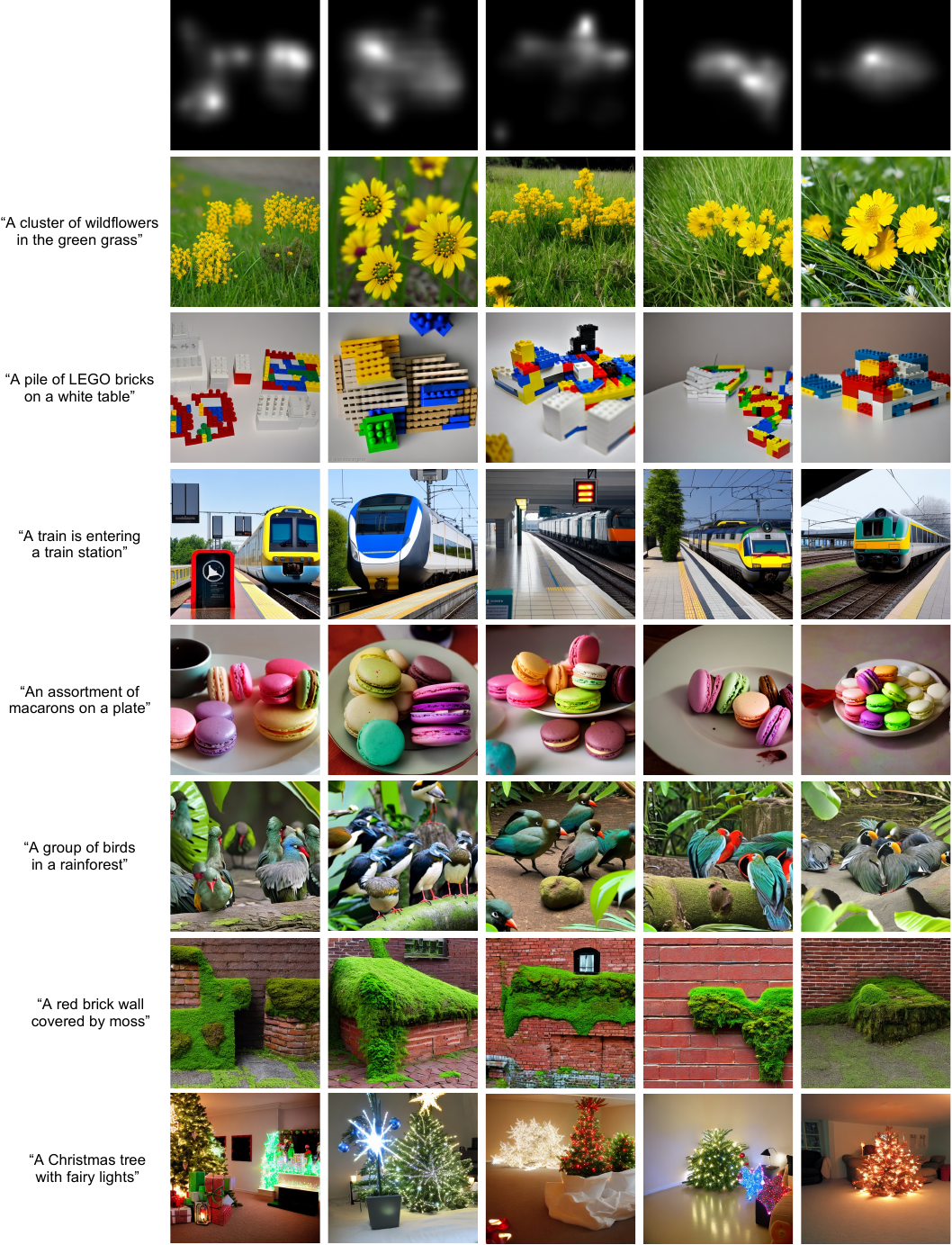}
\Caption{\revise{Saliency-guided image generation with \methodName.}}
{}
\label{fig:ablation-model-image-results}
\end{figure}
\end{document}